\documentclass[sigconf]{acmart}

\usepackage{booktabs} 

\usepackage{graphicx}  
\usepackage{amsmath,url,xcolor} 
\usepackage{subcaption} 

\acmConference[HeteroNAM'18]{International Workshop on
Heterogeneous Networks Analysis and Mining}{February 2018}{Los Angeles, California, USA} 
\acmYear{2018}
\copyrightyear{2018}

\begin{document}
\title{StackSeq2Seq: Dual Encoder Seq2Seq Recurrent Networks}

\author{Alessandro Bay}
\affiliation{%
  \institution{Cortexica Vision Systems}
  \streetaddress{30 Stamford Street}
  \city{London} 
  \state{UK} 
  \postcode{SE1 9LQ}
}
\email{alessandro.bay@cortexica.com}
\author{Biswa Sengupta}
\affiliation{%
  \institution{AI Theory Lab, Noah's Ark}
  \streetaddress{Huawei Research and Development UK}
\city{London, UK} 
}
\email{biswa.sengupta@huawei.com}

\renewcommand{\shortauthors}{A. Bay and B. Sengupta}

\begin{abstract}
A widely studied non-deterministic polynomial time (NP) hard problem lies in finding a route between the two nodes of a graph. Often meta-heuristics algorithms such as $A^{*}$ are employed on graphs with a large number of nodes. Here, we propose a deep recurrent neural network architecture based on the Sequence-2-Sequence (Seq2Seq) model, widely used, for instance in text translation. Particularly, we illustrate that utilising a context vector that has been learned from two different recurrent networks enables increased accuracies in learning the shortest route of a graph. Additionally, we show that one can boost the performance of the Seq2Seq network by smoothing the loss function using a homotopy continuation of the decoder's loss function.
\end{abstract}

%
%



\maketitle

\section{Introduction}

In the intersection of discrete optimization and graph theory lies an age-old problem of finding shortest routes between two nodes of a graph. Many theoretical properties of such shortest path algorithms can be understood by posing them on a graph  \citep{Sedgewick2011}. Such graphs can be an inventory delivery algorithm posed on a road network graph (transportation) to a clustering of similar images and videos (computer vision). Traditionally, such discrete non-deterministic polynomial hard optimisation problems are studied using meta-heuristics algorithms such as the $A^{*}$ algorithm. Other algorithms of notable mention are the Dantzig-Fulkerson-Johnson algorithm \citep{Dantzig1954}, branch-and-cut algorithms \citep{Naddef2001}, neural networks \citep{Ali1993}, etc. Recent work \citep{Bay2017} have proposed that a recurrent neural network (RNN) can learn the adjacency matrix of a graph using as a training set the paths that have been generated using a vanilla $A^*$ algorithm. The shortest path computed by an RNN predicts the following node without using any notion of a neighbourhood i.e., for a graph with $N$ nodes, the search space is unrestricted to $N-1$. 

The primary problem surrounding the recurrent neural network's approximation of the shortest route problem is the difficulty of the network to encode longer sequences. This problem has been partly alleviated with network architectures such as long short-term memory (LSTM, \cite{Hochreiter1997}) and the gated recurrent units (GRU, \cite{Cho2014}). Efforts have also been put towards a Neural Turing Machines \citep{Graves2014} and a differentiable neural computer \citep{Graves2016}  that act as an augmented RNN with a (differentiable) external memory which can selectively be read or written to. 

In this paper, we use the shortest path problem as an empirical example to understand how information about longer sequences can be retained in an RNN. Our testbed -- finding the shortest routes between two points on a graph -- allow us to not only control length dependencies but also the requisite computational complexity of the inference problem. As a path-finding algorithm we formulate a novel recurrent network based on the Sequence-to-Sequence (Seq2Seq, \cite{Sutskever2014}) architecture. Specifically, we show that using context vectors that have been generated by two different recurrent networks can facilitate the decoder to have an increased accuracy in approximating the shortest route estimated by the $A^{*}$ algorithm.  Moreover, our method differs from the Pointer Network (Ptr-Net, \citep{vinyals2015pointer}) as we use two different encoders (one based on LSTM and another based on GRU), while Ptr-Net uses attention from the decoder as a pointer to select a member of the input sequence in the encoder. 

\section{Methods}
\label{sec:methods}

In this section, we describe the data-sets, the procedure for generating the routes for training/test datasets, and the architecture of the dual encoder Seq2Seq network that forms the novel contribution of this paper. All of the calculations were performed on an i7-6800K CPU @ 3.40GHz workstation with 32 GB RAM and a single nVidia GeForce GTX 1080Ti graphics
card.

\subsection{Datasets}

The graph is based on the road network of Minnesota\footnote{\url{https://www.cs.purdue.edu/homes/dgleich/packages/matlab_bgl}}. Each node represents the intersections of roads while the edges represent the road that connects the two points of intersection. Specifically, the graph we considered has 376 nodes and  455 edges, as we constrained the coordinates of the nodes to be in the range $[-97,-94]$ for the longitude and $[46,49]$ for the latitude, instead of the full extent of the graph, i.e., a longitude of $[-97,-89]$ and  a latitude of $[43,49]$, with a total number of 2,642 nodes.

\subsection{Algorithms}

\subsubsection*{The $A^{*}$ meta-heuristics}
\label{sec:astar}

The $A^{*}$ algorithm is a best-first search algorithm wherein it searches amongst all of the possible paths that yield the smallest cost. This cost function is made up of two parts -- particularly, each iteration of the algorithm consists of first evaluating the distance travelled or time expended from the start node to the current node. The second part of the cost function is a heuristic that estimates the cost of the cheapest path from the current node to the goal. Without the heuristic part, this algorithm operationalises the Dijkstra's algorithm \citep{Dijkstra1959}. There are many variants of $A^{*}$; in our experiments, we use the vanilla $A^{*}$ with a heuristic based on the Euclidean distance. Other variants such as Anytime Repairing $A^{*}$ has been shown to give superior performance \citep{Likhachev2004}. 

\begin{figure}[!h]
\centering{
\includegraphics[scale=0.6]{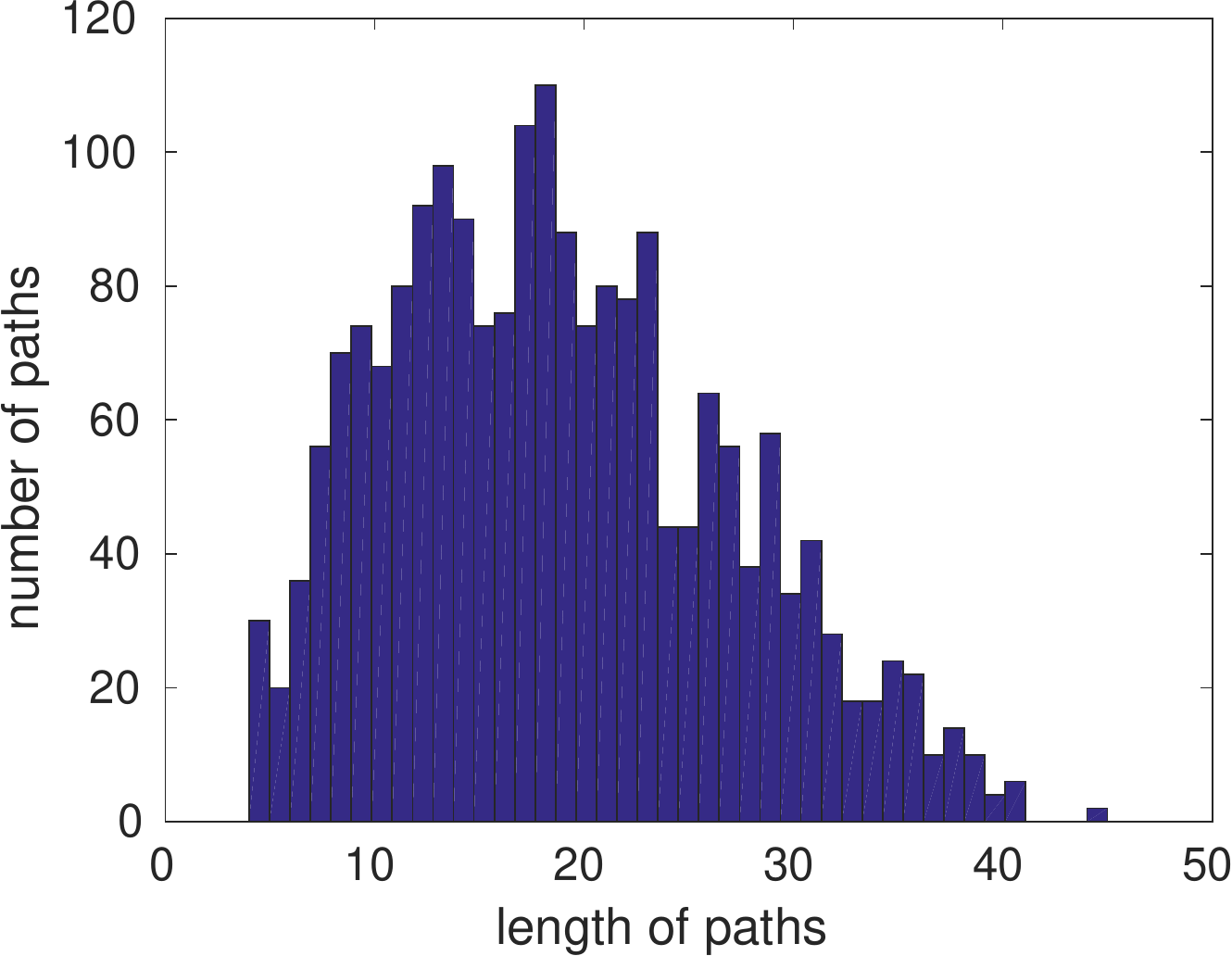}
\caption{\textbf{Distribution of path lengths.} After selecting two nodes uniformly at random, we compute the shortest paths using the $A^*$ algorithm. The average path length is 19 hops.}
\label{fig:histLength}}
\end{figure}

Paths between two randomly selected nodes are calculated using the $A^{*}$ algorithm. On an average, the paths are {19} hops long and follow the distribution represented by the histogram in Figure \ref{fig:histLength}.

\subsubsection*{Recurrent deep networks}
\label{sec:rnn}
We operationalise our network as follows: we start by feeding the source node, that has been embedded in a matrix learned during the training phase, in the encoder. The embedded destination node is then fed as second time step in the RNN encoder. The resulting hidden states of the two encoders are stacked to build the context vector, which maintains the memory trace of the source and the destination. It is then used as an initial hidden state for the decoder. Subsequently, the decoder takes as input the embedding of each node in the training sequence.  During the test phase, instead, the input is the node predicted in the previous time step. 

In order to predict the following node, we use an output layer with a (log-)softmax non-linear function; this gives us a probability distribution that allows us to rank the most probable node among the remaining N-1 nodes (if N is the number of nodes in the graph) corresponding to the subsequent hop. Finally, during the training phase, we use a negative log-likelihood as the cost function to compute the loss, which is backpropagated via backpropagation through time.

In particular, we utilised a variety of Sequence-to-Sequence recurrent neural networks for shortest route path predictions:

\begin{itemize}
\item An LSTM2RNN, where the encoder is modelled by an LSTM, i.e.
\begin{align*}
i(t)&=\textrm{logistic}\bigg(A_i x(t) + B_i h(t-1) + b_i\bigg)\\
j(t)&=\tanh\bigg(A_j x(t) + B_j h(t-1) + b_j\bigg)\\
f(t)&=\textrm{logistic}\bigg(A_f x(t) + B_f h(t-1) + b_f\bigg)\\
o(t)&=\textrm{logistic}\bigg(A_o x(t) + B_o h(t-1) + b_o\bigg)\\
c(t) &= f(t) \odot c(t-1) + i(t) \odot j(t)\\
h(t) &= o(t) \odot \tanh\bigg(c(t)\bigg),
\end{align*}
while the decoder is a vanilla RNN, i.e.
\begin{equation}\label{eq:RNN}
\begin{cases}
h(t) = \tanh(A x(t) + B h(t-1) + b) \\
y(t) = \textrm{logsoftmax}(C h(t) + c)
\end{cases}.
\end{equation}

\item A GRU2RNN, where the encoder is modelled by a GRU, i.e.
\begin{align*}
z(t)&=\textrm{logistic}\bigg(A_z x(t) + B_z h(t-1) + b_z\bigg)\\
r(t)&=\textrm{logistic}\bigg(A_r x(t) + B_r h(t-1) + b_r\bigg)\\
\tilde h(t)&=\tanh\bigg(A_h x(t) + B_h(r(t) \odot h(t-1)) + b_h\bigg)\\
h(t) &= z(t) \odot h(t-1) + (1-z(t)) \odot \tilde{h}(t),
\end{align*}
while the decoder is again a vanilla RNN, as in Equation \eqref{eq:RNN}.

\item A dual context Seq2Seq model, where two different latent representations are learned using two different encoders (one LSTM and one GRU). The context vector takes the form of a stacked latent encoding. In Figure \ref{fig:doubleContext}, we show the two context vectors stacked in a matrix for each path in the training set. For both encoders, their respective matrices are full rank; the stacked context vector is also of full rank. This means that GRU and LSTM encode very different context vectors and it is worth considering  both of them for an accurate encoding.

\begin{figure}[!h]
\centering{
\includegraphics[scale=0.6]{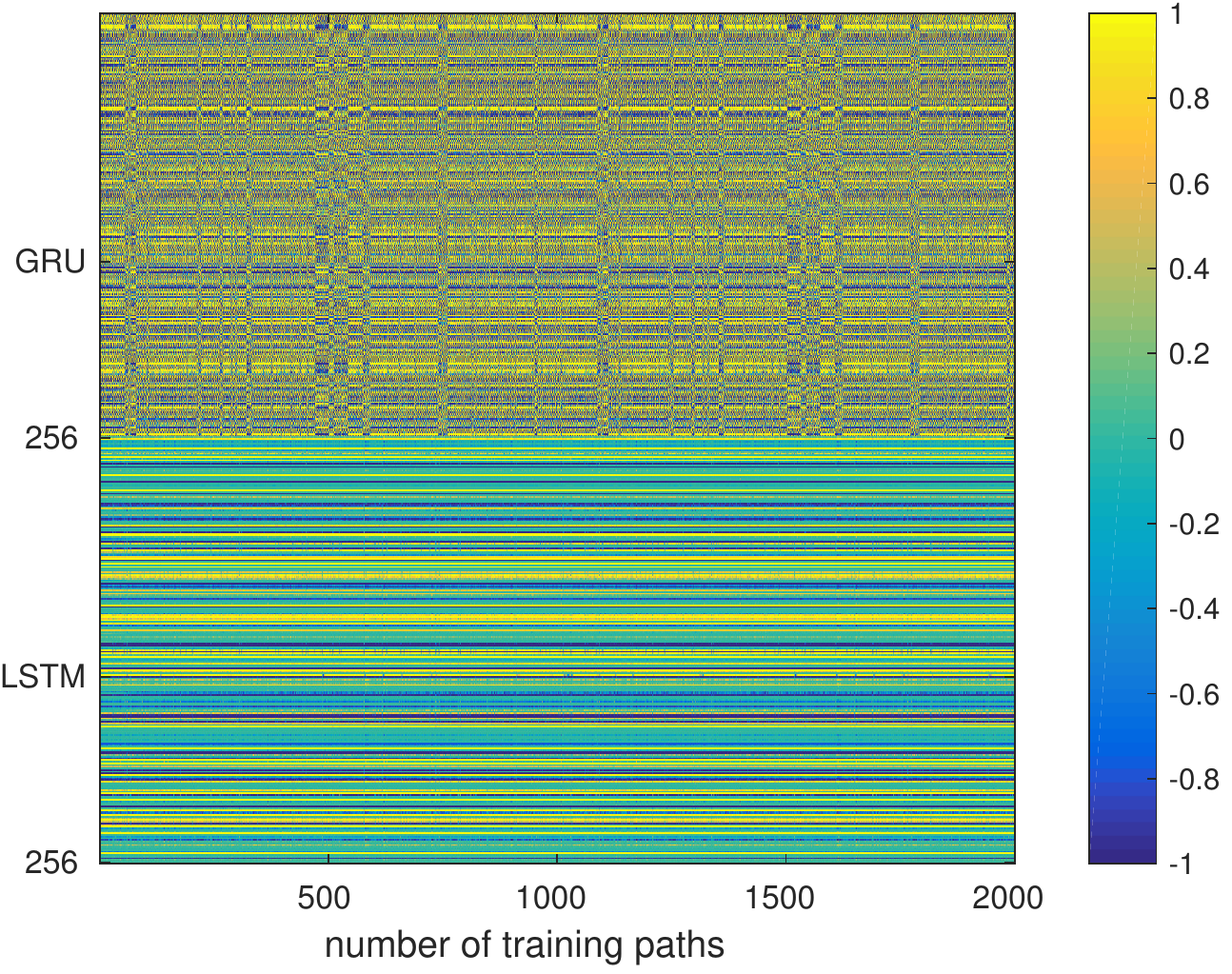}
\caption{\textbf{Context vectors for GRU and LSTM encoders.} Matrices with training context vectors for GRU and LSTM. Their individual and composite rank are full.}
\label{fig:doubleContext}
}
\end{figure}

\item A dual context Seq2Seq model, where two different latent representations are learned using two different encoders (one LSTM and one GRU) and the decoder is represented by a vanilla RNN, trained with homotopy continuation \citep{Vese1999}. This is done by convolving the loss function with a Gaussian kernel -- for more details please refer to \cite{Bay2017}. Our novel contribution lies in extending the framework of  \cite{Mobahi2016} by obtaining an analytic approximation of the log-softmax function. Table \ref{table:diffusedActivations} illustrates the diffused forms of the most popular activation functions.

\begin{table}[!h]
\centering{
\resizebox{1\columnwidth}{!}{
\begin{tabular}{l c c c}
\hline
function & original & diffused \\
\hline
error  & $\textrm{erf}(\alpha x)$ & $\textrm{erf}\left(\frac{\alpha x}{\sqrt{1+2(\alpha\sigma)^2}}\right)$\\
tanh & $\tanh(x)$ & $\tanh\left(\frac{x}{\sqrt{1+\frac{\pi}{2}\sigma^2}}\right)$ \\
sign & $ \begin{cases} +1 \quad \textrm{if } x>0\\ \;\,\,\,0 \quad \textrm{if } x=0 \\ -1 \quad \textrm{if } x<0\end{cases} $ & $\textrm{erf}\left(\frac{x}{\sqrt{2}\sigma}\right)$ \\
relu & $\max(x,0)$ & $\frac{\sigma}{\sqrt{2\pi}}\exp\left(\frac{-x^2}{2\sigma^2}\right)+\frac{1}{2}x\left(1+\textrm{erf}\left(\frac{x}{\sqrt{2}\sigma}\right)\right)$ \\
logsoftmax & $x-\log\bigg(\sum \exp(x)\bigg)$ &  $\bigg(\left(1-\frac{1}{\pi}\right)\exp\left(-\pi\sigma^2\right)+\frac{1}{\pi}\bigg) x - \log(\sum(\exp(x)))$ \\ 
\hline \\
\end{tabular}}
\caption{\textbf{List of diffused forms (Weierstrass transform).} We report the most popular non-linear activation functions along with their diffused form. This is obtained by convolving the function with the heat kernel $K(x,\sigma)$. This table extends the work in \cite{Mobahi2016} by an analytic approximation of the  log softmax function.  For more details please refer to \cite{Bay2017}.}
\label{table:diffusedActivations}}
\end{table}

\end{itemize}

\begin{figure}[!h]
    \centering
    \begin{subfigure}[b]{0.5\textwidth}
        \includegraphics[width=\textwidth]{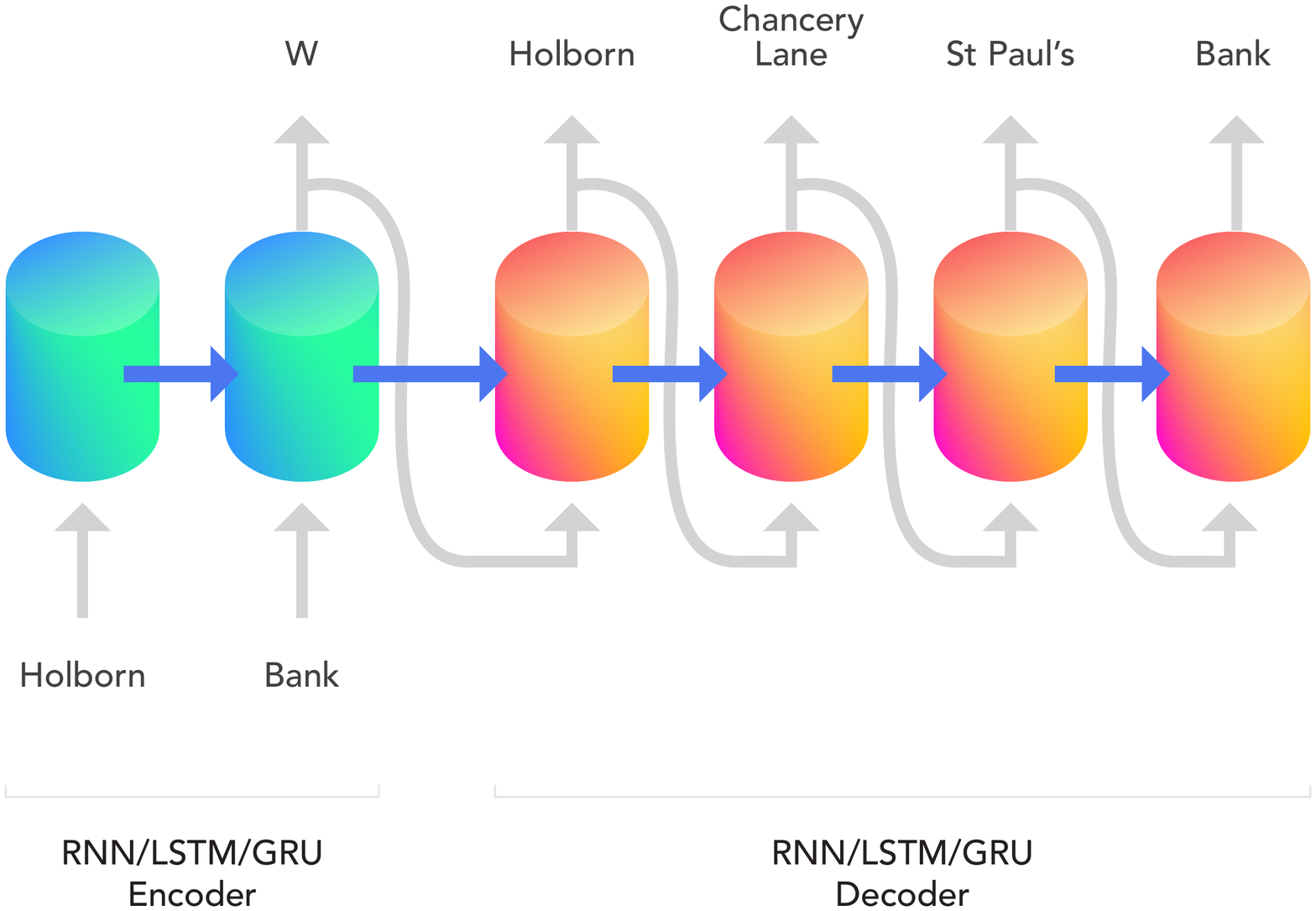}
        \caption{Seq2Seq network}
        \label{fig:gull}
    \end{subfigure}%
      
    \begin{subfigure}[b]{0.5\textwidth}
    \hspace{-2.3cm}
        \includegraphics[width=1.5\textwidth]{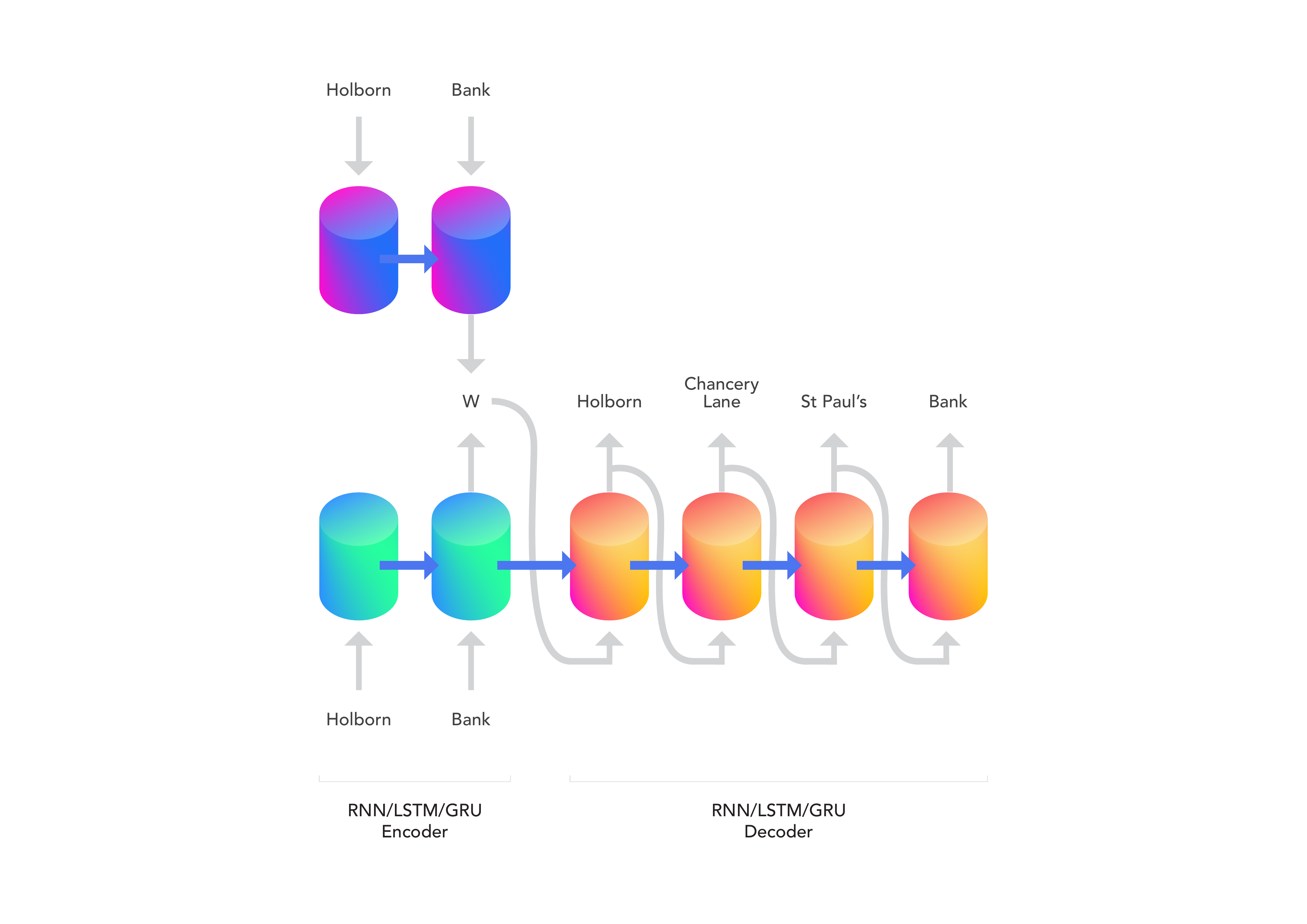}
        \caption{Dual-context Seq2Seq network}
        \label{fig:tiger}
    \end{subfigure}
    
    \caption{\textbf{Dual-context Sequence-to-Sequence architecture for approximating the $A^{*}$ meta-heuristics.} For both networks, the first two modules on the left are the encoder while the last four represent the decoded output, representing the shortest route between Holborn and Bank. The network is trained using shortest route snippets that have been generated using an $A^{*}$ algorithm. $w$ represents the context vector.}
\label{fig:seq2seqGraph}
\end{figure}

For all networks, as shown in Figure \ref{fig:seq2seqGraph}, the input is represented by the [source, destination] tuple, which is encoded in a context vector ($W$) and subsequently decoded into the final sequence to obtain the shortest path connecting the source to the destination. Moreover, during the test phase, we compute two paths, one from the source to the destination node and the other from the destination to the source node, that forms an intersection to result in the shortest path.

\section{Results}
\label{sec:results}

For the graph of Minnesota with 376 nodes and  455 edges, we generated 3,000 shortest routes between two randomly picked nodes using the $A^*$ algorithm. We used these routes as the training set for the Seq2Seq algorithms using a 67-33\% training-test splits. 

For the two encoders involved, we choose a hidden state with 256 units, such that the joint latent dimension of the two neural networks is 512. In our experiments, we compare the standard Seq2Seq with either 256 or 512 hidden units. We run the training for 400 epochs, updating the parameters with an Adam optimisation scheme \citep{Kingma2014}, with parameters $\beta_1=0.9$ and $\beta_2=0.999$, starting from a learning rate equal to $10^{-3}$. On the other hand, for the diffused loss function, we smooth the cost function using a Gaussian kernel of standard deviation $\sigma=\{30, 5, 1, 0.0001\}$. The training iterates converged after 100 epochs for each value of $s$.

The prediction accuracy on the test data-set is reported in Table \ref{table:results}. As we can see, doubling the hidden state dimension marginally increases the percentage of shortest paths ($1\%$) and the successful paths, that are not necessarily the shortest ($0.2\%$ and $1.6\%$ for GRU and LSTM encoders, respectively). Our proposed dual encoder, which is composed by two context vectors of dimension 256 (obtained from the GRU and LSTM encoders, respectively) for a total of 512 components, achieves a $10\%$ improvement on the shortest paths (almost $58\%$). Moreover, if trained with diffusion (homotopy continuation), it turns out to be the best performing algorithm with about $60\%$ of accuracy on the shortest paths and more than $78\%$ on the successful cases. 

This means that our proposed encoder significantly increases the number of the retrieved shortest paths,  thanks to its dual dynamics. It is important to highlight that the dimension of the latent space is proportional to task complexity wherein our experiments demonstrate that doubling the dimension of the context vector, for both LSTM and GRU encoders considered separately, bring only marginal improvements. However, having encodings learned by two different recurrent encoders offer more flexibility.

\begin{table}[!h]
\centering{
\begin{tabular}{l c c c}
\hline 
method & shortest & successful \\
\hline
LSTM2RNN (256) & $47\%$ & $69.5\%$ \\
LSTM2RNN (512) & $48\%$ & $71.1\%$ \\
GRU2RNN (256) & $48\%$ & $73.1\%$ \\
GRU2RNN (512) & $49\%$ & $73.3\%$ \\
dual encoder & $57.7\%$ & $77.1\%$ \\
dual encoder with diffusion& $\textbf{59.6\%}$ & $\textbf{78.3\%}$ \\
\hline \\
\end{tabular}
\caption{\textbf{Results on the Minnesota graph.} Percentage of shortest path and successful paths (that are not necessarily shortest) are shown for a wide-variety of Seq2Seq models, with context vector dimension equal to either 256 or 512. All scores are relative to an $A^{*}$ algorithm, that achieves a shortest path score of 100\%.}
\label{table:results}}
\end{table}

\section{Discussion}
\label{sec:discussion}

Our prime motivation behind the current work was to understand the temporal congruency of a recurrent neural network, not to replace the $A^*$ algorithm with a recurrent neural network. The shortest route problem, approximated by the  $A^*$ algorithm, gives us a flexible way to control the temporal length of a sequence, by simply sampling paths that are longer. Additionally, another handle on the computational complexity of the problem is achieved by increasing or decreasing the size of the graph. Our experiments illustrate that -- (a) recurrent networks have the fidelity to memorize varying lengths of sequences, by learning the adjacency matrix of a given graph and (b) the required latent dimension of the embedding learned by a recurrent network is dependent on the task complexity. However, what remains unclear is how increasing memory capacity of a recurrent network creates/destroys new minimizers  for the neural network i.e., the performance of a high capacity RNN will invariably suffer if ``high quality'' minimizers are difficult to obtain, due to non-convexity of the loss function.

It is clear that using two context vectors instead of one improves the decoder's accuracy in approximating the $A^{*}$ algorithm. What we have proposed in this paper is akin to feature stacking wherein two different sets of features are stacked to increase classification accuracy. Our experiments that control the embedding dimension of the latent context vector (256 or 512) show that the increased number of successful routes produced by the neural network is due to the encoding dynamics, not the encoding dimension. Indeed, a homotopy continuation induced diffusion increases the accuracy by $\approx 2\%$, it still falls short in improving the temporal memory of the encoder.

In future, we foresee using a sequential probabilistic model of the latent context vector that might afford to learn the structure of the sub-route's temporal congruency. 

\subsubsection*{Acknowledgments}

BS is thankful to the Issac Newton Institute for Mathematical Sciences for hosting him during the ``Periodic, Almost-periodic, and Random Operators" workshop.

\bibliographystyle{ACM-Reference-Format}
\bibliography{LSTM_inference_NPhard} 

\end{document}